\documentclass{article}

\usepackage{arxiv}

\usepackage{xspace}
\usepackage{soul}
\usepackage{amsmath,amssymb,amsfonts}
\usepackage{algorithmic}
\usepackage{graphicx}
\usepackage{textcomp}
\usepackage{graphicx}
\usepackage[table,xcdraw]{xcolor}
\usepackage{multirow}
\usepackage{ulem}
\usepackage{booktabs}
\usepackage{siunitx}
\usepackage{placeins}
\usepackage{subcaption}
\usepackage{cite}

\usepackage{xcolor}

\newcommand\extrafootertext[1]{%
    \bgroup
    \renewcommand\thefootnote{\fnsymbol{footnote}}%
    \renewcommand\thempfootnote{\fnsymbol{mpfootnote}}%
    \footnotetext[0]{#1}%
    \egroup
}

\def\BibTeX{{\rm B\kern-.05em{\sc i\kern-.025em b}\kern-.08em
    T\kern-.1667em\lower.7ex\hbox{E}\kern-.125emX}}
\begin{document}
\title{Towards Scalable IoT Deployment for Visual Anomaly Detection via Efficient Compression}

\author{
    Arianna Stropeni \\ 
    University of Padova, Italy \\ 
    \texttt{arianna.stropeni@studenti.unipd.it} \\ \And 
    Francesco Borsatti \\ University of Padova, Italy \\ 
    \texttt{francesco.borsatti.1@phd.unipd.it} \\ \And
    Manuel Barusco \\ University of Padova, Italy \\ 
    \texttt{manuel.barusco@phd.unipd.it} \\ \And
    Davide Dalle Pezze \\ University of Padova, Italy \\ 
    \texttt{davide.dallepezze@unipd.it} \\ \And
    Marco Fabris \\ University of Padova, Italy \\ 
    \texttt{marco.fabris.1@unipd.it} \\ \And
    Gian Antonio Susto \\ University of Padova, Italy \\
    \texttt{gianantonio.susto@unipd.it} \\
}

\maketitle

\begin{abstract}
Visual Anomaly Detection (VAD) is a key task in industrial settings, where minimizing operational costs is essential. Deploying deep learning models within Internet of Things (IoT) environments introduces specific challenges due to limited computational power and bandwidth of edge devices. This study investigates how to perform VAD effectively under such constraints by leveraging compact, efficient processing strategies. We evaluate several data compression techniques, examining the tradeoff between system latency and detection accuracy. Experiments on the MVTec AD benchmark demonstrate that significant compression can be achieved with minimal loss in anomaly detection performance compared to uncompressed data.
Current results show up to 80\% reduction in end-to-end inference time, including edge processing, transmission, and server computation.
\end{abstract}


\section{Introduction}
Visual anomaly detection (VAD) is essential in industrial automation to detect defective parts by image-based inspection. Traditional VAD systems operate in centralized environments with high performance computing and reliable data acquisition infrastructures \cite{YANG2022471}. However, with the rise of the Internet of Things (IoT), this paradigm is evolving. IoT enables distributed sensing for real-time asset monitoring in remote environments. Industrial IoT (IIoT) networks connect production sites, supporting flexible monitoring architectures \cite{CHATTERJEE2022100568,FahimAnomaly2019}. These changes introduce new constraints for VAD systems, as edge devices have limited resources and bandwidth, requiring the restructuring of VAD pipelines. This has spurred the development of lightweight, communication-efficient frameworks based on unsupervised deep learning and distributed processing \cite{GangeAnomaly2019}.

VAD research ranges from computer vision to deep learning, as documented in surveys \cite{YANG2022471}. Studies have explored anomaly detection under resource constraints \cite{barusco2024paste}. Edge-assisted systems reduce latency in industrial monitoring \cite{CHATTERJEE2022100568}, while \cite{FahimAnomaly2019} proposes low-latency VAD for smart cities. Research has also explored edge computing integration \cite{GangeAnomaly2019}, traffic flow analysis \cite{fabris2025efficient}, health monitoring systems \cite{Frigo2015heartrate} and data-driven predictive control \cite{breschi2023impact}.

In IIoT, research has focused on communication resource allocation \cite{RoseroHybrid2023} and federated learning for distributed detection \cite{GuoUnsupervised2020}. Studies have examined edge-cloud architecture trade-offs \cite{mohindru2020review}. Optimization in VAD includes performance-aware frameworks \cite{Ceschini2017Optimization} and adaptive models \cite{GAURAV2025503}; whereas, investigations on effective multiobjective methods aim to reduce latency and memory usage while maintaining accuracy \cite{DURAIRAJ2024111919}.

While a standard taxonomy of VAD methods is lacking, recent studies \cite{liu2023simplenet, quan2025omni} propose a classification into three categories: reconstruction-, synthesizing-, and embedding-based approaches. Reconstruction methods assume anomalies are poorly reconstructed due to their absence in the training data (e.g., visual sensor networks in \cite{VAROTTO2022105096}); using autoencoders, GANs, diffusion models, and Transformers to reconstruct samples, identifying anomalies via pixel-wise or SSIM differences. However, overfitting or visually similar anomalies can lead methods to underperform. Synthesizing methods generate artificial anomalies on normal images, such as CutPaste \cite{li2021cutpaste} and DRÆM \cite{zavrtanik2021draem}, but sometimes struggle to capture real-world defect characteristics. 
Finally, embedding methods use pretrained models to represent normal data distributions; e.g. PaDiM \cite{defard2021padim}, PatchCore \cite{roth2021total}, FastFlow \cite{yu2021fastflow}. 
These methods achieve high accuracy but can be resource-intensive and sensitive to domain shifts.

Nonetheless, to the best of our knowledge, existing literature lacks solutions that jointly optimize VAD models and communication strategies for IoT.
This paper proposes a lightweight, embedding-based VAD framework for bandwidth-constrained IIoT environments, with efficient data transmission and interpretable anomaly localization. Our contributions are as follows:

\begin{itemize}
	
\item Resource-aware VAD pipeline for distributed edge-server deployment, optimizing detection performance, model footprint, and communication overhead for efficient anomaly detection in IoT environments.

\item Test two strategies: edge image compression VS tiny CNN feature extraction with compression, reducing computational and communication resources while preserving VAD performance.
We test compression methods like WebP for images, random projection and product quantization for feature encoding.

\item We validate the framework using the MVTec AD dataset and MobileNetV2 as the edge feature extractor, showing that our method retains detection accuracy close to uncompressed data, suitable for real-world IIoT deployments.
\end{itemize}

The remainder of the manuscript unfolds as follows. Sec.~\ref{sec:background} provides background on visual anomaly detection and formalizes the problem in an IoT context. Sec.~\ref{sec:method} presents the proposed resource-aware VAD framework, describing both the system architecture and the methods for compressing features and images. Sec.~\ref{sec:results} reports experimental results on the MVTec AD dataset, evaluating trade-offs between latency, bandwidth, and detection accuracy under various deployment scenarios. Finally, Sec.~\ref{sec:conclusions} concludes the paper and outlines directions for future research.

\begin{figure*}[!th]
    \centering
    \includegraphics[width=0.6\textwidth]{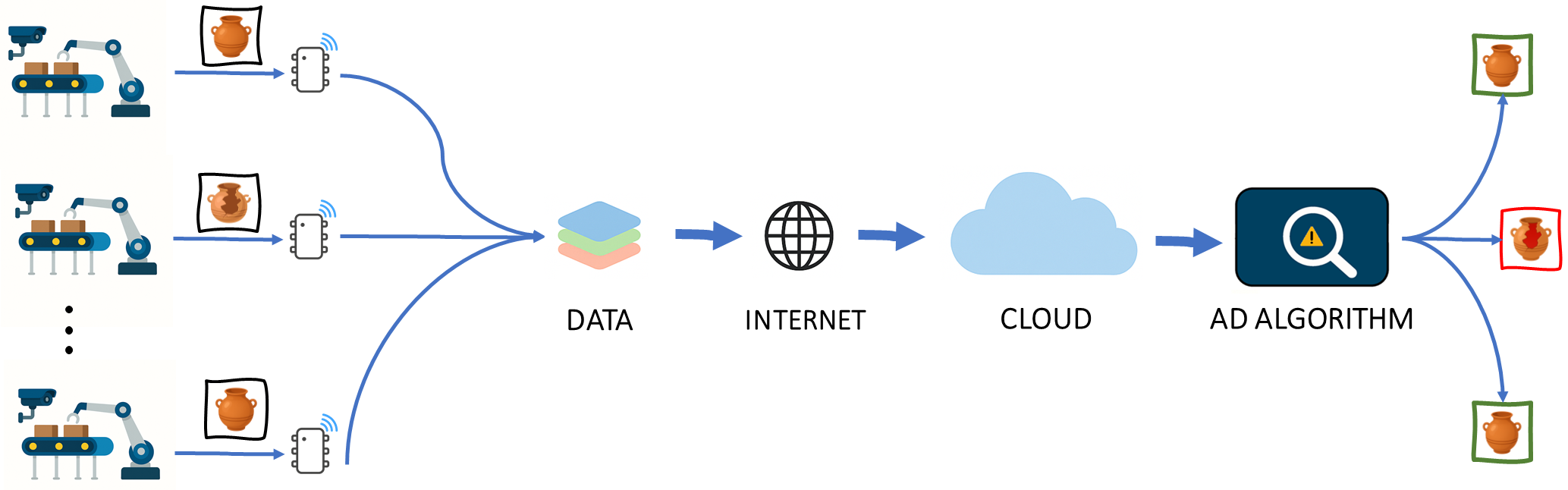}
    \caption{ Overview of the VAD pipeline for IoT scenarios: edge devices acquire raw asset images and run compression techniques or apply feature extraction. Data are then sent across a bandwidth-limited IIoT channel to a central server. An anomaly detector (PatchCore in this study) is used to identify anomalous samples and localize defects.  }
    \label{fig:MIIC_dataset}
\end{figure*}

\section{Background and problem formulation} \label{sec:background}
In the sequel we provide the necessary background on VAD to formalize the optimization problem under analysis.

\subsection{Mathematical preliminaries}

We denote the set of CNN feature vectors as $\mathcal{F} = \{f_1, \dots, f_N\} \subset \mathbb{R}^d$, where each $f_i$ is obtained by concatenating features from $L$ intermediate layers at the same spatial location. A random subset $\mathcal{F}_s \subset \mathcal{F}$ is selected such that $|\mathcal{F}_s| = \alpha N$ with $\alpha \in (0,1)$ denoting the sampling ratio.
For an input image $x$, the feature map at layer $\ell$ is denoted by $\phi^{(\ell)}(x) \in \mathbb{R}^{C_\ell \times H_\ell \times W_\ell}$. 
The operator $\text{flatten} : \mathbb{R}^{C \times H \times W} \to \mathbb{R}^{CHW}$ maps a 3D feature tensor to a vector. 
The global image embedding is  constructed as
\[
v = \text{concat}(f^{(1)}, \ldots, f^{(m)}), \quad m>1,
\]
where $\text{concat}$ denotes vector concatenation. 
The vector $v \in \mathbb{R}^d$ is compressed using product quantization, denoted
\[
\text{PQ} : \mathbb{R}^d \to \mathbb{R}^{d'} \quad (d' \ll d),
\]
yielding the quantized representation 
\begin{equation}\label{eq:quantrepr}
    \hat{v} = \text{PQ}(v).
\end{equation}

\subsection{Problem formulation}\label{sec:problem}

Let $\mathcal{X} = \{x_i\}_{i=1}^n$ denote a set of input images acquired by edge devices in an industrial IoT environment. Each image $x_i \in \mathbb{R}^{H \times W \times 3}$ must be analyzed for anomalies using an unsupervised VAD model. Also, let $\mathcal{E}$ represent the set of edge devices, each subject to constraints on computational capacity and communication bandwidth.
Our goal is to learn a detection function
\[
p : \mathbb{R}^{H \times W \times 3} \to \mathbb{R}, \quad p(x) \in [0,1],
\]
where $p(x)$ represents an anomaly score, with higher values indicating greater likelihood of abnormality.

The distributed system nature, whose pipeline is composed of two stages $p(x) = f(g(x))$, imposes two constraints:
\begin{itemize}
    \item \textbf{Computation Constraint:} Each edge device $e \in \mathcal{E}$ has limited computational capacity, modeled as an upper bound $C_e$ on admissible operations/ model size.
    \item \textbf{Communication Constraint:} For each device $e$, the transmitted data $z_i = g(x_i)$ must satisfy $\|z_i\| \leq B_e$, where $g$ is a compression or feature extraction function and $B_e$ is the communication budget.
\end{itemize}

We aim to jointly design:
\begin{itemize}
    \item a transmission function $g : \mathbb{R}^{H \times W \times 3} \to \mathbb{R}^d$ (either image compression or feature extraction),
    \item and a centralized detection model $f : \mathbb{R}^d \to [0,1]$,
\end{itemize}
such that the expected anomaly detection performance is maximized:
\[
\max_{f, g} \; \mathbb{E}_{x \sim \mathcal{X}}[\text{AUC}(f(g(x)))]
\]
subject to the constraints:
\[
\text{cost}(g; C_e) \leq C_e, \quad \|g(x)\| \leq B_e, \quad \forall x \in \mathcal{X}, \; e \in \mathcal{E}.
\]

\section{Proposed method}\label{sec:method}

In this work we address the task of unsupervised VAD within an IoT-enabled industrial setting, where resource constraints on edge devices impose limitations on both computational capacity and communication bandwidth. In the following, we present effective solutions to the objective defined in Sec.~\ref{sec:problem}, aiming to design a detection pipeline that operates across distributed nodes while minimizing data transmission costs and preserving detection accuracy.

\begin{figure}[htbp]
    \centering
    \includegraphics[width=\linewidth]{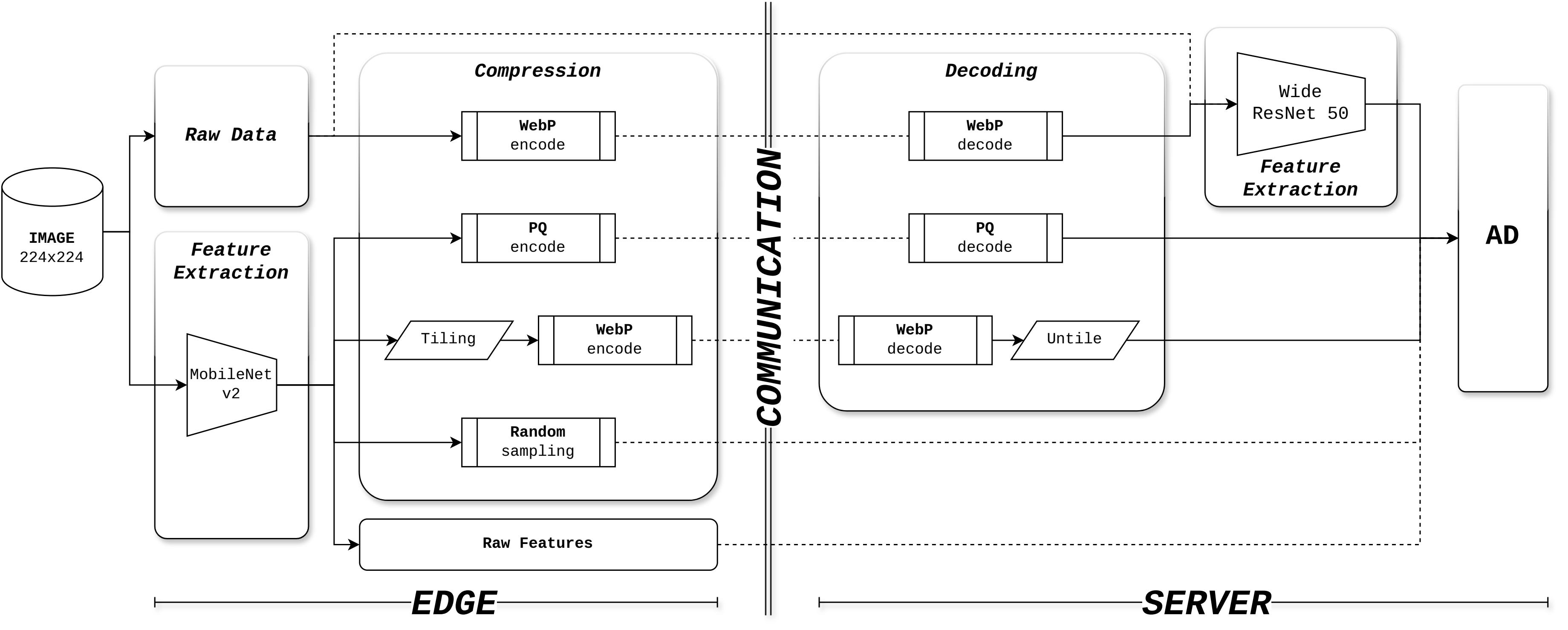}
    \caption{Illustration of the visual anomaly detection system architectures evaluated in this study. Different compression strategies are applied at the edge to raw data and features before transmission to the server for decoding and feature extraction.}
    \label{fig:diagram}
\end{figure}

The proposed pipeline is illustrated in Fig.~\ref{fig:diagram}. On the edge, the raw image data is first subjected to feature extraction via a lightweight convolutional neural network (CNN), such as MobileNetV2. Depending on the selected deployment strategy, this step is followed by either image compression or feature compression.

Once the compressed data reaches the server, the corresponding decoding operations are performed. For image compression, WebP decoding is applied, while for feature compression, PQ decoding is used. After decoding, the image or feature data undergoes a second feature extraction phase, typically using a more complex CNN like WideResNet50, to obtain embeddings suitable for anomaly detection. Finally, the processed data is passed through the anomaly detection (AD) module, which generates the anomaly score, indicating whether the input is anomalous or not.

This modular approach allows for flexibility in dealing with the trade-off between computational load on the edge and communication bandwidth requirements. In the following subsections, we will delve into each stage of this pipeline, examining the advantages and challenges associated with the various techniques employed.

\subsection{Deployment architecture}
\label{sec:deployed_architecture}

To address computational heterogeneity and communication constraints in IoT environments, we propose a two-part distributed architecture, shown in Fig.~\ref{fig:MIIC_dataset}. The core VAD model is hosted on a centralized server, while edge devices are responsible for either image acquisition and compression, or feature extraction (and optional feature compression). Two deployment strategies are considered:

1) \textbf{Image Compression at the Edge}: The device acquires an image, compresses it using standard codecs such as WebP, and transmits the compressed image to the server. This approach requires minimal memory overhead and implementation complexity, though encoding may be slower on CPU-bound devices.

2) \textbf{Feature Extraction at the Edge}: After image acquisition, the device extracts intermediate features using a lightweight CNN (e.g., MobileNetV2). The resulting representation is then transmitted to the server. Although this strategy increases the memory footprint due to the model parameters, it reduces server-side processing and can be advantageous in fog computing scenarios. In this scenario, the resulting feature vectors are still high-dimensional and would pose a communication bottleneck if transmitted directly. To mitigate this, the pipeline includes a feature compression stage, detailed in Section~\ref{sec:feature_compression}, which reduces the transmission size while preserving detection-relevant information.

These strategies are designed as alternatives, enabling flexible deployment depending on the hardware profile of the edge device and the bandwidth constraints of the network. The choice between them directly impacts inference latency, communication load, and overall system scalability.

\subsection{VAD modeling and training}
\label{sec:VAD_modeling_training}

The VAD model is trained solely on normal samples. At inference time, it produces two outputs: (i) an image-level anomaly score and (ii) a pixel-level anomaly map, which enhances interpretability by highlighting anomalous regions. To achieve this, the model adopts feature-based methods for visual anomaly detection, which offer a favorable trade-off between detection performance and computational cost. These methods enable localized detection and compact feature representations suitable for transmission and storage.

Our previous work \cite{barusco2024paste} demonstrated the viability of feature-based approaches in constrained settings. However, not all tasks can be performed at the edge due to computational limitations. IoT enables the deployment of more powerful anomaly detection models on the server, where computational and memory constraints are less restrictive.
We thus consider two system variants based on the distribution of computation between edge and server. In the first, feature extraction is performed at the edge using a lightweight CNN, specifically a MobileNetV2 model pretrained on ImageNet. The extracted features are transmitted to the server, where anomaly detection is carried out. In the second, the edge device applies the WebP codec to compress the image, which is then sent to the server. The server performs feature extraction using a WideResNet50 model.

The key distinction in modeling lies in the dimensionality and expressiveness of the features: MobileNetV2 generates lower-dimensional representations than WideResNet50. Since both configurations employ memory bank–based models (e.g., PatchCore or PaDiM) residing on the server, the memory bank derived from MobileNetV2 features is more compact, enabling more efficient inference and greater scalability across many connected devices. Conversely, the use of WideResNet50 combined with WebP compression yields higher anomaly detection performance due to richer feature representations.

The experimental results presented in Sec.~\ref{sec:results} will provide a detailed discussion of the architectural trade-offs among bandwidth, scalability, and detection accuracy.

\subsection{Feature Compression Techniques}
\label{sec:feature_compression}

For the scenario 2) detailed in \ref{sec:deployed_architecture}, involving on-device feature extraction, additional feature compression is applied to reduce communication overhead. We evaluate several strategies:

\textbf{Random Sampling.} As adopted in several feature-based methods such as PaDiM \cite{defard2021padim}, random sampling of feature maps significantly reduces dimensionality with minimal performance degradation, even at sampling factors of 0.25.

Let $\mathcal{F} = \{f_1, f_2, \dots, f_N\}$ denote the set of feature vectors extracted from $N$ spatial locations across the output of $L$ selected layers of a CNN (e.g., MobileNetV2). Each $f_i \in \mathbb{R}^d$ is the concatenation of features across the $L$ layers at the corresponding spatial location. Random sampling selects a subset $\mathcal{F}_s \subset \mathcal{F}$ such that $|\mathcal{F}_s| = \alpha N$, where $\alpha \in (0,1)$ is the sampling ratio.

\textbf{Product Quantization (PQ).} We apply product quantization \cite{jegou2010product} to compress the image embedding formed by concatenating the feature maps from a set of selected intermediate layers, referred to as feature extraction layers.

Let $L = \{1, 2, 3\}$ be the indices of the chosen feature extraction layers of the CNN. For a given input image $x$, let $\phi^{(\ell)}(x) \in \mathbb{R}^{C_\ell \times H_\ell \times W_\ell}$ denote the feature map output at layer $\ell \in L$.
Each feature map is flattened into a vector:
\[
f^{(\ell)} = \text{flatten}(\phi^{(\ell)}(x)) \in \mathbb{R}^{C_\ell \cdot H_\ell \cdot W_\ell}.
\]

The image-level embedding is defined as the concatenation of all flattened feature maps from the selected layers:
\[
v = \text{concat}(f^{(1)}, f^{(2)}, f^{(3)}) \in \mathbb{R}^d, \quad \text{with }  d = \sum\limits_{\ell \in L} C_\ell H_\ell W_\ell.
\]
This vector $v$ is then compressed via product quantization in \eqref{eq:quantrepr} 
and the quantized representation $\hat{v}$ is transmitted to the server for anomaly detection.

\textbf{Image-Based Compression of Feature Maps.} As an alternative to PQ, we explore the use of a standard codec, specifically WebP, applied directly to intermediate feature maps as proposed by \cite{choi2018deep} for feature compression in object detection tasks.

The combination of lightweight CNNs and lossy but efficient compression mechanisms provides a viable solution for real-world IIoT applications, enabling high compression rates while preserving the interpretability and effectiveness of the anomaly detection process.

\begin{table*}[htbp]
\centering
\caption{Comparison of anomaly detection performance metrics.}
\label{tab:metrics}
\renewcommand{\arraystretch}{1.2}
\resizebox{\columnwidth}{!}{%
\begin{tabular}{llccccccc}
\toprule
\textbf{Metric} & \textbf{Class} & \textbf{Original} & \textbf{Raw Features} & \textbf{WebP} & \textbf{RS} & \textbf{PQ} & \textbf{RS+WebP} & \textbf{RS+PQ} \\
\midrule
\multirow{4}{*}{F1 (pxl)} 
& Objects         & 0.606 & 0.581 & 0.568 & 0.502 & 0.447 & 0.386 & 0.439 \\
& Textures        & 0.498 & 0.407 & 0.528 & 0.367 & 0.256 & 0.239 & 0.241 \\
& Overall         & 0.570 & 0.523 & 0.555 & 0.457 & 0.383 & 0.337 & 0.373 \\
& $\Delta$ Overall (\%) & 0.00\% & $-8.20$\% & $-2.62$\% & $-19.79$\% & $-32.76$\% & $-40.87$\% & $-34.60$\% \\
\midrule
\multirow{4}{*}{ROC (img)} 
& Objects         & 0.979 & 0.977 & 0.976 & 0.950 & 0.877 & 0.877 & 0.870 \\
& Textures        & 0.995 & 0.956 & 0.992 & 0.927 & 0.700 & 0.771 & 0.661 \\
& Overall         & 0.984 & 0.970 & 0.982 & 0.942 & 0.818 & 0.842 & 0.800 \\
& $\Delta$ Overall (\%) & 0.00\% & $-1.43$\% & $-0.28$\% & $-4.29$\% & $-16.94$\% & $-14.47$\% & $-18.73$\% \\
\bottomrule
\end{tabular}
}
\end{table*}

\section{Experimental results}\label{sec:results}

In this section, we evaluate the proposed VAD framework on IoT scenarios, describing the setup, transmission strategies, detection performance, and latency to highlight trade-offs between accuracy and efficiency.

\subsection{Setting and data set}

To evaluate the feasibility of deploying our method on resource-constrained edge devices, we compare inference times obtained on a high-performance server with estimated times for a typical edge device. All experiments were conducted using a single core of an AMD Ryzen Threadripper PRO 5995WX (128 cores) @ 2.7\,GHz. To approximate the performance on an edge device, such as a Raspberry Pi-class processor operating at approximately 900\,MHz, we scale the measured inference times by a factor of 3, corresponding to the ratio between the server and edge CPU clock frequencies (2.7\,GHz / 0.9\,GHz). This provides a coarse estimation of the latency expected on edge hardware under CPU-only execution. 

For communication-related experiments, we assume a constant uplink bandwidth of 100\,KBps per edge device, representing a typical constrained wireless link in practical IoT deployments. Communication latency and protocol overhead are assumed to be zero and are excluded from the scope of this analysis.

The MVTec AD Dataset \cite{bergmann2019mvtec} is an extensive repository of images designed specifically for evaluating AD algorithms. 
The MVTec AD includes ten object categories: Bottle, Cable, Capsule, Hazelnut, Transistor, Metal Nut, Pill, Screw, Toothbrush, and Zipper and five texture categories: Carpet, Grid, Leather, Tile, and Wood.
Each category is provided with defect-free training images and annotated test images featuring a wide variety of anomalies (e.g., scratches, dents, contaminations) along with pixel ground-truth masks.

\subsection{Examined scenarios}

In this work, we evaluate various data transmission strategies within an IoT-based visual anomaly detection (VAD) framework. Each scenario reflects a different balance between compression rate, computational cost, and anomaly detection performance.

All configurations are based on PatchCore \cite{roth2021total} as the anomaly detection method, and all input images from the MVTec AD dataset \cite{bergmann2019mvtec} are uniformly resized to $224 \times 224$. This resolution is standard in the VAD literature (e.g., PaDiM (\cite{defard2021padim}), CFA \cite{lee2022cfa}) and is compatible with the input sizes of the MobileNetV2 and WideResNet backbones used in our experiments.

\begin{itemize}
    \item Feature Extraction with Random Sampling (0.25): The edge device extracts features using MobileNetV2, retains 25\% of the patches via random sampling, and sends them to the server.

    \item Product Quantization (PQ): MobileNetV2-extracted features are compressed with PQ before transmission, and decoded on the server.

    \item Random Sampling (0.50) + PQ: Combines random sampling of MobileNetV2-extracted feature maps (50\% retention) with PQ compression. 

    \item Random Sampling (0.50) + WebP Compression on Feature Maps: Feature maps from each layer are tiled and compressed as an image using WebP as described by \cite{choi2018deep}. After transmission, the compressed image is converted back to feature tensors.

    \item WebP Compression (Quality 80) + Server-Side Feature Extraction: The image is WebP-compressed at the edge, then decoded and processed on the server using WideResNet.

    \item Raw Image Transmission ($224 \times 224$): The downsampled image is sent directly to the server, where WideResNet performs feature extraction and anomaly detection.

    \item Raw Feature Transmission: Features are extracted on-device via MobileNetV2 and sent uncompressed to the server.
\end{itemize}

\subsection{Anomaly detection performance}
We evaluate anomaly detection performance at both the image and pixel levels. Image-level accuracy is assessed using the Area Under the ROC Curve (ROC AUC), measuring the ability to classify whether an image contains any anomaly. Pixel-level performance is measured using the F1-score, which evaluates the accuracy of the predicted anomaly heatmap against ground-truth segmentation masks.

Table~\ref{tab:metrics} summarizes the image-level and pixel-level results, aggregated by object and texture categories, as well as overall performance. On the pixel-level metric, the WebP-based approach shows only a 2\% drop in F1-score compared to the uncompressed baseline, while the MobileNetV2 edge feature extractor results in an 8\% drop. All remaining methods degrade significantly, with losses between 20\% and 40\%, indicating a substantial reduction in localization accuracy.

For image-level anomaly detection, WebP again performs best among the lightweight methods, with only a 0.3\% reduction in ROC AUC relative to the baseline. Other methods exhibit larger losses ranging from 16\% to 20\%, highlighting their reduced classification reliability.

\begin{figure}[htbp]
    \centering
\includegraphics[trim={0.3cm 0.2cm 0.2cm 0.3cm},clip,width=0.85\linewidth]{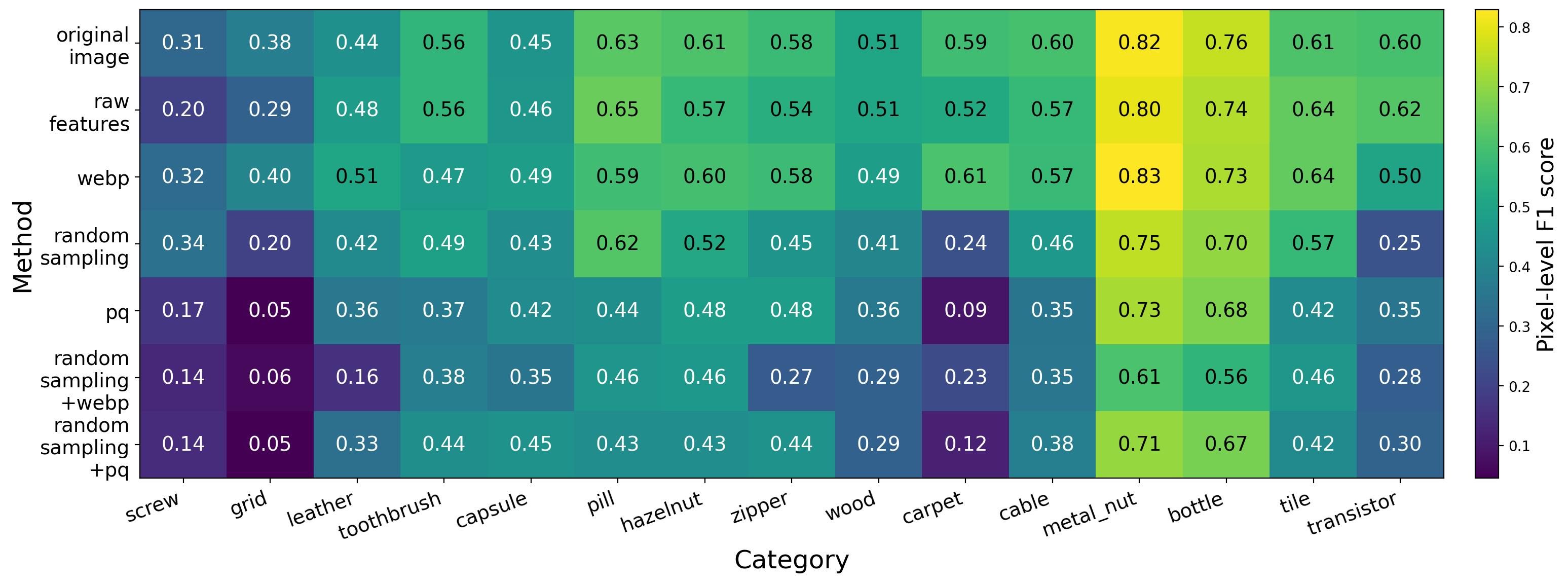}
    \caption{Pixel-level F1-scores across all MVTec categories.}
    \label{fig:heatmap_f1}
\end{figure}

Overall, the WebP compression strategy preserves sufficient visual information to maintain high segmentation and classification performance across all MVTec categories, demonstrating its effectiveness and robustness. In contrast, the edge-feature approach using MobileNetV2 yields less consistent results, primarily due to the limited representational capacity of the lightweight feature extractor. These findings, summarized by Fig.~\ref{fig:heatmap_f1}, confirm that deferring feature extraction to the server —after high-quality compression— offers a better trade-off between communication cost and detection accuracy in resource-constrained IoT deployments. However, edge-side feature extraction remains advantageous in scenarios where server-side scalability and load balancing are critical, as it eliminates the need for server-side feature extraction and reduces the size of the memory bank, as discussed in Sec.~\ref{sec:VAD_modeling_training} and previous work~\cite{barusco2024paste}.

\subsection{Inference latency}

The end-to-end inference latency is evaluated by accounting for edge-side feature extraction and optional compression, transmission over the fixed-bandwidth channel, server-side decoding, optional feature extraction, and anomaly detection using a KNN-based approach as in PatchCore. A detailed breakdown of total times for each scenario is reported in Table~\ref{tab:comparison_total_times}.

Table~\ref{tab:edge_server_times} reports the latency breakdown for the main components of the pipeline: edge-side computation (either feature extraction, compression, or both) and server-side processing (which includes decompression, optional feature extraction, and anomaly detection inference).

\begin{table}[t]
\centering
\caption{Inference times per image (ms).}
\label{tab:edge_server_times}
\begin{tabular}{lcclccc}
\toprule
\textbf{Method} &
  \multicolumn{2}{c}{\textbf{Edge}} &
  \multicolumn{3}{c}{\textbf{Server}} &
  \textbf{Tot.} \\
\cmidrule(r){2-3} \cmidrule(lr){4-6}
 &
  \begin{tabular}[c]{@{}c@{}}Feature\\ extracttion\end{tabular} &
  Encoding &
  Decoding &
  \begin{tabular}[c]{@{}c@{}}Feature\\ extraction\end{tabular} &
  AD &
  \\
\midrule
Original        & --     & --     & --       & 0.74  & 109.4 & 110.1 \\
Features        & 15.0   & --     & --       & --    & 105.7 & 120.7 \\
RS.25           & 46.1   & --     & --       & --    & 102.0 & 148.1 \\
WebP            & --     & 9.1    & 0.1      & 0.6   & 109.5 & 119.3 \\
PQ              & 30.0   & 60.0   & 0.15     & --    & 106.8 & 197.0 \\
RS.5+PQ         & 30.0   & 17.7   & 0.10     & --    & 103.8 & 151.6 \\
RS.5+WebP       & 30.0   & 9.3    & 2.9      & --    & 103.2 & 145.4 \\
\bottomrule
\end{tabular}
\end{table}

As shown in Table~\ref{tab:comparison_total_times}, the method using WebP compression stands out in terms of end-to-end latency, thanks to the high efficiency of the codec. Additionally, hybrid strategies such as Random Sampling with Product Quantization (RS+PQ) and Random Sampling combined with WebP also achieve low latency, despite the additional cost of using a CNN-based feature extractor at the edge.

In contrast, transmitting raw features is prohibitively expensive under constrained bandwidth conditions, resulting in significantly higher total inference times.

The results assume a fixed transmission bandwidth of 100~kB/s, which is a realistic figure for Industrial IoT (IIoT) applications. Total time accounts for edge processing, transmission delay (Tx Time), and server-side computation. Payload refers to the data volume sent to the server. Although we plan to explore varying bandwidths in future work, the current results already demonstrate substantial efficiency gains, achieving up to 80\% reduction in end-to-end inference time.

\begin{table}[htb]
\centering
\caption{Image processing times. Payload is the size of the data sent to the server. Tx Time is the transmission time over a 100~kB/s channel. Total Time includes edge, transmission, and server processing. $\Delta$Time indicates total time change vs. baseline (Original). Lower is better for all metrics.}
\label{tab:comparison_total_times}
\setlength{\tabcolsep}{4pt}
\begin{tabular}{lcccc}
\toprule
\textbf{Method} & \textbf{Payload} & \textbf{Tx Time} & \textbf{Total Time} & \textbf{$\Delta$Time} \\
               & {[kB]}           & {[s]}            & {[s]}               & {[vs. Orig]}          \\
\midrule
Original        & 60   & 0.60  & 0.71  & 0\%     \\
Raw Features    & 382  & 3.82  & 3.94  & +455\%  \\
WebP            & 2    & 0.02  & 0.14  & --80\%  \\
RS              & 95   & 0.95  & 1.10  & +55\%   \\
PQ              & 4    & 0.04  & 0.24  & --67\%  \\
RS + WebP       & 2    & 0.02  & 0.17  & --76\%  \\
RS + PQ         & 2    & 0.02  & 0.17  & --77\%  \\
\bottomrule
\end{tabular}
\end{table}

\subsection{Performance vs. Efficiency Trade-off}

\begin{figure}[htbp]
    \centering
    \includegraphics[width=0.6\linewidth]{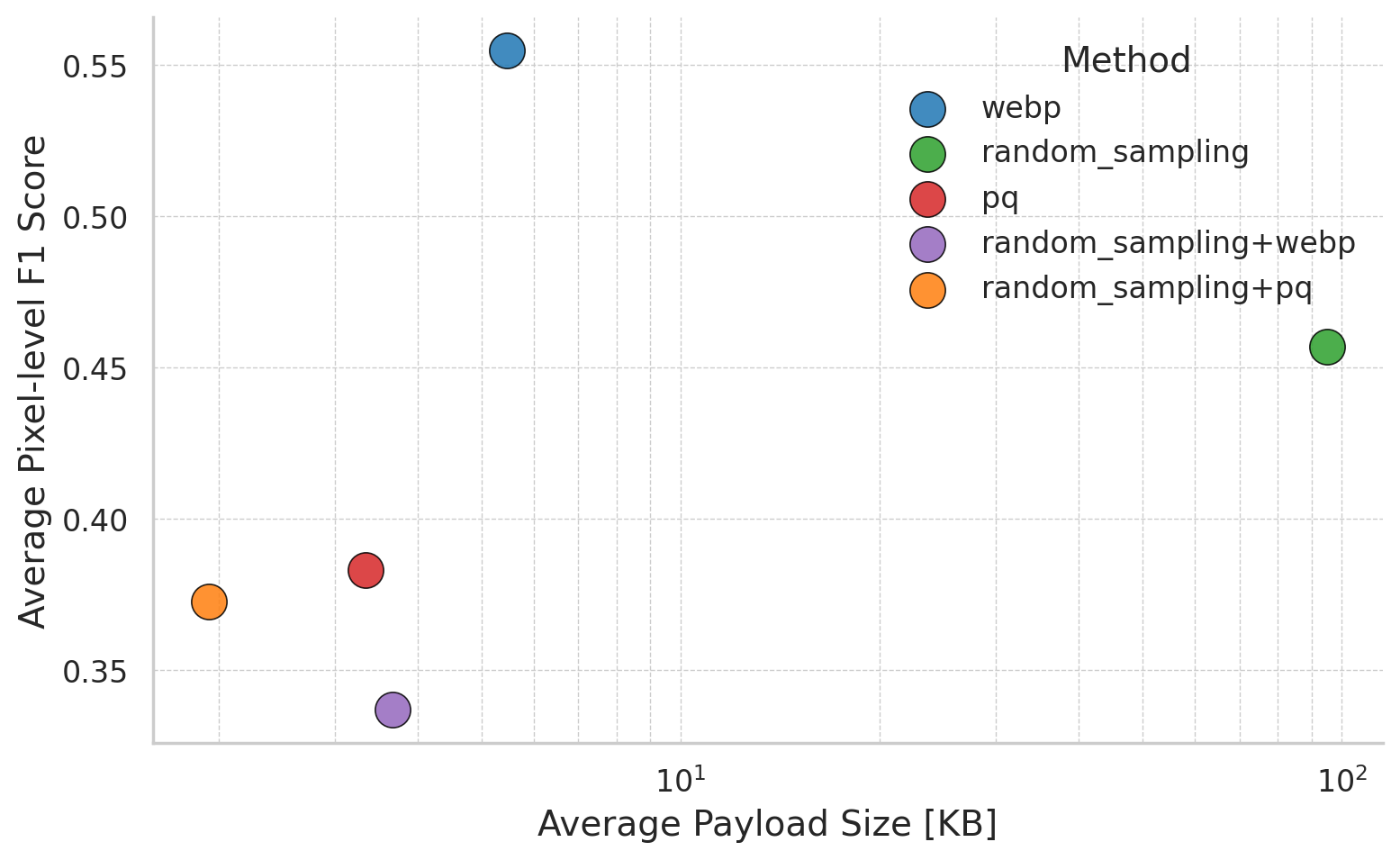}
    \caption{Trade-off between pixel-level F1-score and average payload size for each method.}
    \label{fig:scatter}
\end{figure}

To better understand the trade-off between anomaly detection performance and communication efficiency, we combine the evaluation of pixel-level F1-score and transmitted payload size in a single scatter plot, shown in Figure~\ref{fig:scatter}.

Each point represents a method, positioned according to its pixel-level F1-score and the average payload size sent over the network. As expected, the WebP method lies closest to the ideal top-left region (high performance, low size), confirming it as the most effective compromise between accuracy and efficiency.

\section{Conclusions and future directions}\label{sec:conclusions}

We have proposed an efficient VAD framework for IoT scenarios, included in our open-source library MoViAD\footnote{https://github.com/AMCO-UniPD/moviad} \cite{barusco2025moviadmodularvisualanomaly}.
Testing both image and feature compression strategies shows that communication overhead can be reduced while maintaining high detection accuracy. Validated on the MVTec AD dataset using a lightweight CNN and PatchCore, our approach matches the performance of uncompressed pipelines with lower latency and memory usage, making it viable for industrial deployment.
Future work will explore dynamic adaptation strategies that adjust compression based on real-time factors like bandwidth and computational load. We also plan to investigate learning-based feature compression and extend the framework to support online model updates and real-time anomaly localization across heterogeneous edge-server systems. Another promising direction involves embedding security mechanisms directly into the framework, aiming to enhance the resilience of heterogeneous edge-server networks \cite{fabris2022secure,fabris2023robustness}. 

\bibliographystyle{IEEEtran}
\bibliography{references}

\end{document}